\documentclass{article}

\usepackage{arxiv}

\usepackage[utf8]{inputenc} 
\usepackage[T1]{fontenc}    
\usepackage{hyperref}       
\usepackage{url}            
\usepackage{booktabs}       
\usepackage{amsfonts}       
\usepackage{nicefrac}       
\usepackage{microtype}      
\usepackage{graphicx}
\usepackage[numbers]{natbib}
\usepackage{doi}
\usepackage{float}

\title{FatNet: High Resolution Kernels for Classification Using Fully Convolutional Optical Neural Networks}

\author{ \href{https://orcid.org/0000-0002-0359-0830}{\includegraphics[scale=0.06]{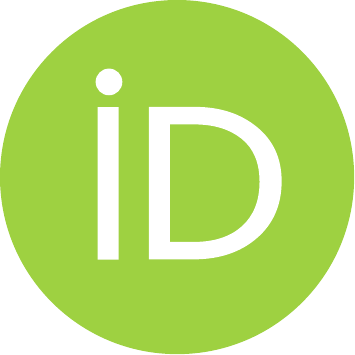}\hspace{1mm}Riad Ibadulla*} and \href{https://orcid.org/0000-0001-8037-1685}{\includegraphics[scale=0.06]{orcid.pdf}\hspace{1mm}Thomas M. Chen} and 	\href{https://orcid.org/0000-0002-9466-2018}{\includegraphics[scale=0.06]{orcid.pdf}\hspace{1mm}Constantino Carlos Reyes-Aldasoro}
    \\ \\
	Department of Computer Science, School of Science \& Technology\\
	City, University of London\\
	Northampton Square, London, EC1V 0HB \\ \\
	Correspondence*: \texttt{riad.ibadulla@city.ac.uk} \\
}

\hypersetup{
pdftitle={FatNet: High Resolution Kernels for Classification Using Fully Convolutional Optical Neural Networks},
pdfsubject={cs.cv},
pdfauthor={Riad Ibadulla, Thomas Chen, Constantino Carlos Reyes-Aldasoro},
pdfkeywords={optical neural networks, high resolution, convolutional neural networks},
}

\begin{document}
\maketitle

\begin{abstract}
This paper describes the transformation of a traditional in-silico classification network into an optical fully convolutional neural network with high-resolution feature maps and kernels. When using the free-space 4f system to accelerate the inference speed of neural networks, higher resolutions of feature maps and kernels can be used without the loss in frame rate. We present FatNet for the classification of images, which is more compatible with free-space acceleration than standard convolutional classifiers. It neglects the standard combination of convolutional feature extraction and classifier dense layers by performing both in one fully convolutional network. This approach takes full advantage of the parallelism in the 4f free-space system and performs fewer conversions between electronics and optics by reducing the number of channels and increasing the resolution, making the network faster in optics than off-the-shelf networks. To demonstrate the capabilities of FatNet, it trained with the CIFAR100 dataset on GPU and the simulator of the 4f system, then compared the results against ResNet-18. The results show 8.2 times fewer convolution operations at the cost of only 6\% lower accuracy compared to the original network. These are promising results for the approach of training deep learning with high-resolution kernels in the direction towards the upcoming optics era.
\end{abstract}


\section{Introduction}

One of the major problems of the modern deep learning approach is the speed of training and inference of architectures that have a very large number of parameters to train. Computer vision, which can involve a large number of images with very slight differences, is considered to be one of the most complex problem areas for AI. Within the deep learning approaches, convolutional neural networks (CNNs) have become a standard approach for various computer vision problems. Recently CNNs have been successfully applied to image classification~\cite{krizhevsky_imagenet_2017}, object detection~\cite{redmon_you_2016}, localization~\cite{tompson_efficient_2015}, and segmentation~\cite{ronneberger_u-net_2015} among many other applications. CNNs are suitable for computer vision tasks because neurons in CNNs are not connected to every single neuron of the next layer as in fully-connected networks, but only to the pixels of their receptive field. This approach reduces the number of trainable parameters, which accelerates the inference and makes the neural network more immune to overfitting. Although CNNs are computationally less expensive than fully connected neural networks, accelerating CNNs is also an important task, especially with the ever growing number of images that are captured.

There are many techniques to accelerate deep learning training, e.g., using shallow networks, pruning redundant weights or using lower quantization levels~\cite{rastegari_xnor-net_2016}. In addition, hardware accelerators can be used to speed up the training and inference of neural networks, for example, application-specific integrated circuits (ASICs) which can outperform standard CPUs and GPUs \cite{sunny_survey_2021}. Large tech companies are actively working on their AI accelerators, such as Google's TPU \cite{jouppi_-datacenter_2017}, Intel's Loihi \cite{davies_loihi_2018}, and IBM's TrueNorth \cite{debole_truenorth_2019}. Unfortunately, these accelerators are starting to face limitations in the post Moore's law era since the computational power of the processors is not improving at the same pace as before \cite{waldrop_chips_2016}. 

Optical processors are an interesting alternative to processing data with silicon chips. Optical computing uses photons of light as the information carrier instead of electrons or electric current for data processing~\cite{li_fundamentals_2018}. Since Moore's law does not affect optical computing, optical accelerators can be used for deep learning offering advantages such as the high bandwidth of the light beam, high speed, zero resistance, lower energy consumption, and immunity to overheating~\cite{lin_all-optical_2018}. There are two main approaches to optical neural networks: free-space using spatial light modulators (SLM)~\cite{li_channel_2020,chang_hybrid_2018} or silicon photonics approach using Mach-Zehnder interferometers (MZI)~\cite{shen_deep_2017,hughes_training_2018}. Unlike the silicon photonics approach, free-space optics uses wireless light propagation through a medium which can be air, outer space or vacuum. Although the silicon photonics approach is faster as its clock speed can reach several GHz, it is inferior to the free-space system in parallelism~\cite{sui_review_2020}. 

In the context of optics, one of the most important concepts is the focal distance (f), i.e., the distance at which all beams would concentrate after passing through a convex lens. If the beam continues to travel, the beam would have returned after another distance (2f) to the same position as before the lens. 

This research is focused on the 4f free-space approach as described in Li et al.~\cite{li_channel_2020}, which takes advantage of the parallelism of free-space optics. The 4f free-space optical system can be used to perform convolution operations faster than traditional electronic processors.

The \textit{Fourier transform} is a well-known mathematical operation that decomposes the signal into the fundamental sinusoids in the frequency domain that when combined form the original function~\cite{bracewell_fourier_2000}. A Fourier transform is initially defined over one dimension, and can be extended to two or more dimensions, which is the case for applications in computer vision. One way to do a 2D Fourier transform would be to apply a 1D transform to one dimension of the data (like the rows of an image) and then another 1D transform of the other dimension (the columns)~\cite{gaskill_linear_1978, bracewell_fourier_2000}. The computational complexity of this process increases with the dimensions of the data, and even with fast methods like the fast Fourier transform~\cite{cooley_algorithm_1965}, transforming large data can take considerable resources with complexity in the order of $O(n^2log(n))$, where $n^2$ is the number of pixels of an image~\cite{colburn_optical_2019}. On the other hand, performing a 2D Fourier transform in free-space optics can be easily achieved by passing the light through the convex lens, where the light has to travel only two focal distances (f) from the lens~\cite{jutamulia_fourier_2002}. 

The well known convolution theorem states that if $f(x)$ has the Fourier transform $F(s)$ and $g(x)$ has the Fourier transform $G(s)$, then the convolution of $f(x)$ and $g(x)$ has the Fourier transform $F(s)G(s)$~\cite{bracewell_fourier_2000}. This means that the Fourier transform of a convolution of two signals is the pointwise product of their Fourier transforms. Taking the convolution theorem into account, the convolution of two signals can be represented as the inverse Fourier transform of the pointwise product of their Fourier transforms. The 4f correlator is based on the Fourier transform properties of the convex lenses~\cite{culshaw_fourier_2020} and performs the convolution operation based on the convolution theorem. Any convex lens projects a Fourier transform of the input object located on the front focal plane onto the back focal plane~\cite{culshaw_fourier_2020}, where it can be pointwise multiplied by the kernel in the Fourier domain. After passing through the second lens, it can be converted back into the space domain. The system is called 4f because the light in the 4f system travels four focal distances of the lens. Hence the 4f approach can accelerate convolutional neural networks by performing the Fourier transforms at the speed of light. The parallelism advantage of the 4f system comes from the theoretically infinite resolution that is bounded in reality by the resolution of the modulators and the camera. 

\begin{figure}[H]
\centering
\includegraphics[width=13 cm]{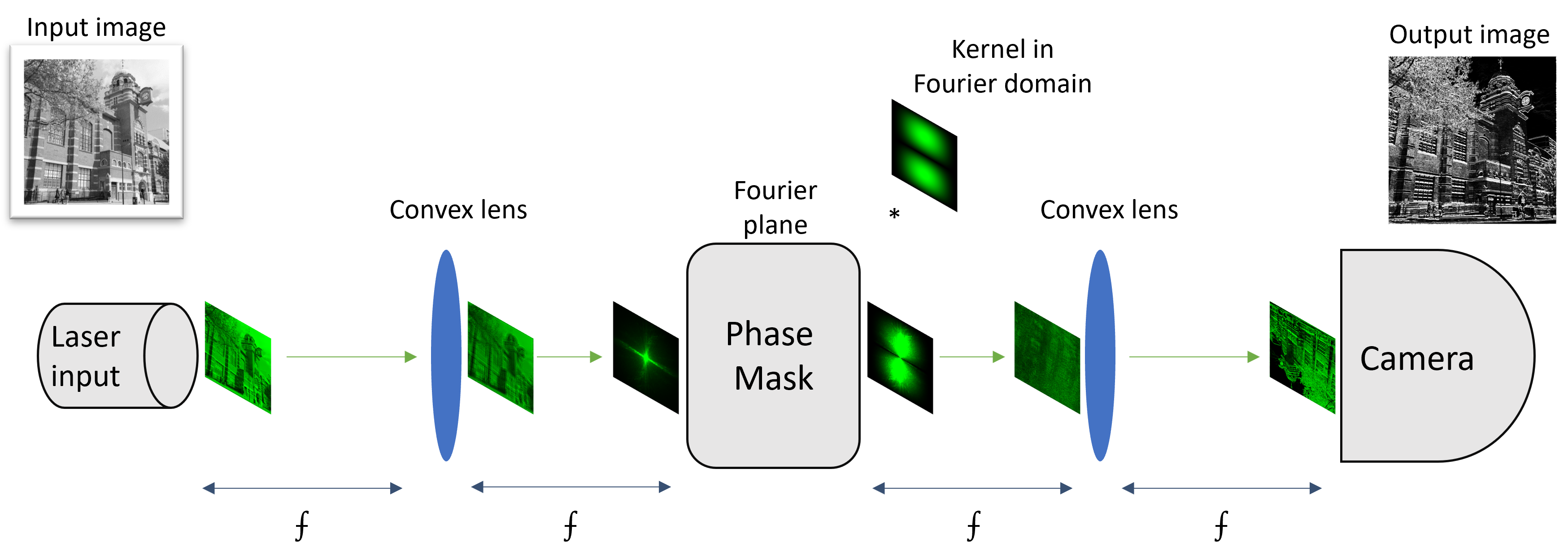}
\caption{Graphical representation of the 4f system performing the convolution operation, consisting of the input plane(laser), the convex lens, Fourier plane(modulator or phase mask), another convex lens and the camera separated from each other in one focal distance of the lens. When light passes through the lens, it forms a 2D Fourier transform on the Fourier plane where it can be multiplied by the kernel in the frequency domain. The light then passes through the second lens, which converts it back into the space domain, where the output is read by the camera.
\label{fig1}}

\end{figure}

The first optical convolution technique with the 4f system was described by Weaver and Goodman~\cite{weaver_technique_1966} in 1966. It was not used for the acceleration of neural networks until neural networks started gaining popularity in the 21st century~\cite{jutamulia_overview_1996}. A standard 4f optical system consists of an input source, two convex lenses, two light modulators and a sensor (see Figure~\ref{fig1}). The input source is the laser emitting the light modulated right in the beginning with the input image by altering the light intensity. The modulated light passes through the first convex lens after travelling the focal distance of the lens and is projected onto the focal plane, where the Fourier transform of the input is formed. On the focal plane using another modulator, the input is element-wise multiplied with the kernel in the frequency domain. After the multiplication in the Fourier domain, the light is passed through the second lens to perform the inverse Fourier transform and captured by the camera or the array of photodetectors. In some cases, instead of the modulator, the fixed phase mask is used to perform the multiplication in the Fourier plane, as demonstrated in Chang et al. \cite{chang_hybrid_2018}. 

The 4f system is used in combination with the electronic compound, called an optical-electronic hybrid system~\cite{chang_hybrid_2018}. This system is used only for inference, and training is performed using the simulator. The networks were trained using the simulator, and the phase mask of the trained kernels of the first layer was fabricated. Those fabricated kernels were used only for the inference of the pre-trained first layer. Hence the inference of the first convolutional layer is computed optically, the output of the electronic network then fitted into the electronic portion of the network. This allows the multiplication to be performed passively, without energy consumption and with zero latency. It also allowed the high speed-up since the first layer of the network is usually the heaviest due to the high resolutions, which the free-space optics can handle for free. Since the optical-electronic hybrid system has used kernel tiling, this system can perform several convolution operations of the first layer in parallel without losing frame rate and power. However, a passive architecture like this lacks flexibility and can only be used with one set of kernels, meaning it can not be reused for all network layers. This is why, in this paper, we are considering only active 4f architecture, allowing the device to perform all convolutional layers of the network by altering the kernel on the Fourier plane. 

Unlike standard neural networks, optical neural networks involve various bottlenecks and constraints. Since the read-out camera captures the intensity of light which is the square of the amplitude, it is impossible to perform the computations with negative values. One of the ways of getting around the problem can be the non-negative constraint, which can significantly affect the accuracy. One way around this constraint is called pseudo-negativity, which can address the restriction to positive values by doubling the number of filters  \cite{chang_hybrid_2018}. This method only uses positive values for the kernel by labelling half of the kernels as positive and the other half negative. After the read-out, the results of 'positive' convolutions are subtracted from the 'negative' results, thus providing the correct convolution operation's outcome. Another bottleneck is the resolution of the modulator and the camera. Although modern cameras can capture up to 4K resolution, this limit does not allow many channels to be tiled and high-resolution feature maps to be used in combination with tiling. 

Despite the many advantages of 4f systems, they have not been popular among the modern AI accelerators. The main problem lies in the very slow cameras and light modulators used in the system. However, there is a possibility to get the acceleration using parallelism and performing several convolution operations simultaneously. For example, Li et al.~\cite{li_channel_2020}  proposed kernel, channel and mixed tiling approaches to better utilise the resolution of the 4f system. Their approaches enable all convolution operations for specific output channels and sum them using one inference through the 4f system. A technique used by \cite{li_channel_2020} applies zero-padding to the input channels and tiles them into one big input block while their corresponding kernels are tiled in the same manner forming a kernel block. This method takes advantage of the massive parallelism of free-space optics. It performs all convolutions of each output channel of the convolutional layer, including the channel summation via one optical inference. By optically convolving the input block and the kernel, the summation of all convolutions of those particular input channels with output channels appears in the middle of the output tensor. This significantly reduces the number of conversions between optics and electronics. That is why it is essential to use the high-resolution capabilities of the 4f system. 

Tiny kernel resolutions became one of the nuances of building CNN architectures. Kernel sizes of 3x3 or 5x5 became the standard for CNNs~\cite{krizhevsky_imagenet_2017}. Although sometimes in ResNet architectures, a large kernel size can be seen in the first layer of the networks  \cite{he_deep_2016}. Theoretically, having a small kernel size has a range of advantages. Reducing the kernel's size reduces the number of trainable parameters, making the network more immune to overfitting and increasing the speed of training. Modern neural networks are all trained on electronic systems, whose training time depends on the number of parameters. This led to the development of architectures with a very small resolution of kernels.
For the same reason, all classifier architectures were developed in the cone shape, where the image is pooled down at every layer, making it faster for the CPU/GPU process. However, this works entirely different for optical neural networks. Due to the nature of free-space optics, the use of large kernels in 4f system-based neural networks will not affect the inference time. Unfortunately, almost all the attempts to train the convolutional neural networks on the 4f system were based on the standard convolutional cone-shaped architectures. 

To overcome the underutilization of the 4f system, we propose {\bf FatNet}, which takes advantage of the high-resolution capabilities of the 4f system by using fewer channels and larger input/kernel resolution in CNNs. Since the resolution does not affect the speed of inference in the 4f system, increasing the resolution and reducing the number of channels makes the network perform fewer convolution operations. This means fewer translations from optics to electronics since the main bottleneck of the system is based on optics-electronics conversions. Our approach does not require pooling between most layers, which speeds up the inference even more at an insignificant cost in loss of accuracy.

\section{Materials}
\begin{figure}[h]
    \centering
    \includegraphics[scale=0.3]{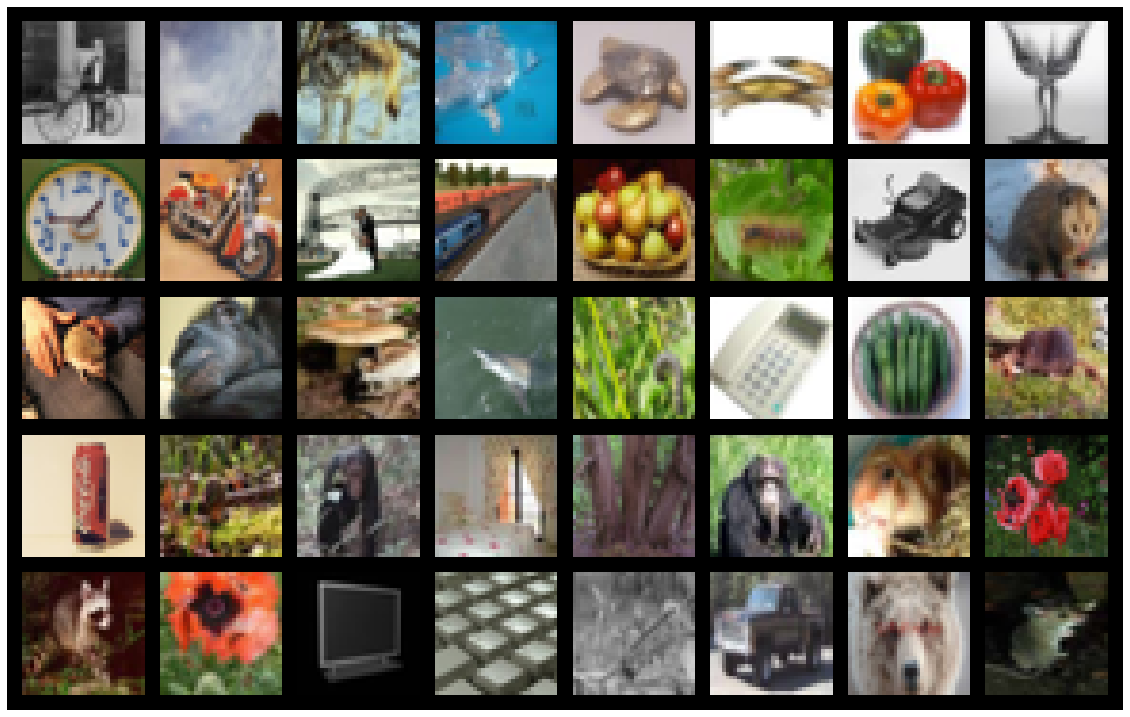}
    \caption{Illustration of CIFAR-100 dataset examples. CIFAR-100 contains tiny images of 100 classes, with a resolution of 32x32.}
    \label{fig:cifar100}
\end{figure}

We trained our network with the CIFAR-100 dataset (see Figure~\ref{fig:cifar100}) and chose the ResNet-18 as the backbone network.
The CIFAR-100 (Canadian Institute For Advanced Research) dataset consists of 60,000 images of 32x32 resolution. It is split into 20 superclasses sub-grouped into 100 classes, with 600 images per class~\cite{krizhevsky_learning_2009}. Only 50,000 images are used for training, and the other 10,000 data samples are in the test set. The similarity of classes under the same superclass in CIFAR-100 makes it harder to train. 

Shah et al. \cite{shah_deep_2016} managed to train CIFAR-100 using different ResNet models, including their variation, where ELU (Exponential Linear Unit)~\cite{clevert_fast_2016} was used as an activation function. Their test error on standard ResNet-101 achieved 27.23\%. For this reason, we decided to use residual networks in our experiments.

ResNet-18 is a CNN, one of five networks introduced in He et al. \cite{he_deep_2016} for the ImageNet dataset~\cite{deng_imagenet_2009}. The distinguishing feature of these networks from others is the residual connections between layers. Formally He et al.\cite{he_deep_2016} noted the blocks of the networks as:
\begin{equation}
    y = F(x,\{W_i\})+x
\end{equation}
where $x$ and $y$ are the input and the output of the residual block, $F(x,\{W_i\})$ represents the building block of the residual layer, which can contain one or several weight layers. 

Residual connections are the connections which skip one or more layers. In ResNet, those connections perform identity mapping, and the outputs of these connections are added to the output of the stacked layers. This configuration allows the use of deeper networks by avoiding vanishing/exploding gradient problems. 

PyTorch was used to train all our networks \cite{paszke_pytorch_2019}. PyTorch is an open source machine learning framework originally developed by Meta AI. We used PyTorch for its flexibility and easiness in creating custom neural network layers. One example can be the simulation of our optical layer, which we also built using PyTorch. PyTorch was also used by Miscuglio~\cite{miscuglio_massively_2020} to precisely simulate the actual 4f system.

\section{Methods}
Nearly all classifier CNNs are cone-shaped and use either strides or pooling layers to reduce the resolution of the feature map~\cite{peng_large_2017}. This architecture has several advantages. The main advantage is the training speed since the network gets simpler after each feature extraction and ends up with very low-resolution feature maps, which are flattened and passed to the fully connected layers for further classification. Moreover, the 3x3 kernels have become the de-facto choice since the smaller kernels mean less trainable parameters, which speeds up the training and makes the network more immune to overfitting. However, this kind of structure became a standard only due to the dominance of electronic computing. Unlike in electronics, having larger resolutions for inputs and kernels in the 4f system do not affect the speed of inference, which means that the exploration of the new architectures compatible with optics becomes essential. Our approach is called FatNet, due to its barrel shaped structure and most of the kernels being the same resolution as the feature maps.

\begin{figure}[H]
    \centering
    \begin{tabular}{cc}
    (a)     &          \includegraphics[width=17cm]{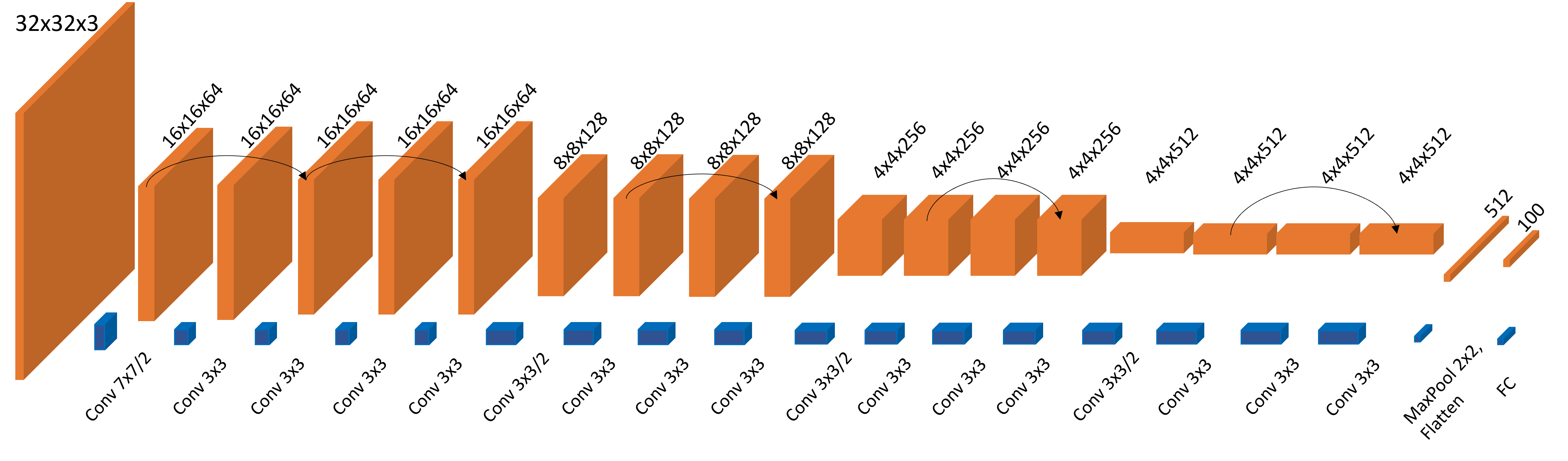}\\
    (b)     &         \includegraphics[width=17cm]{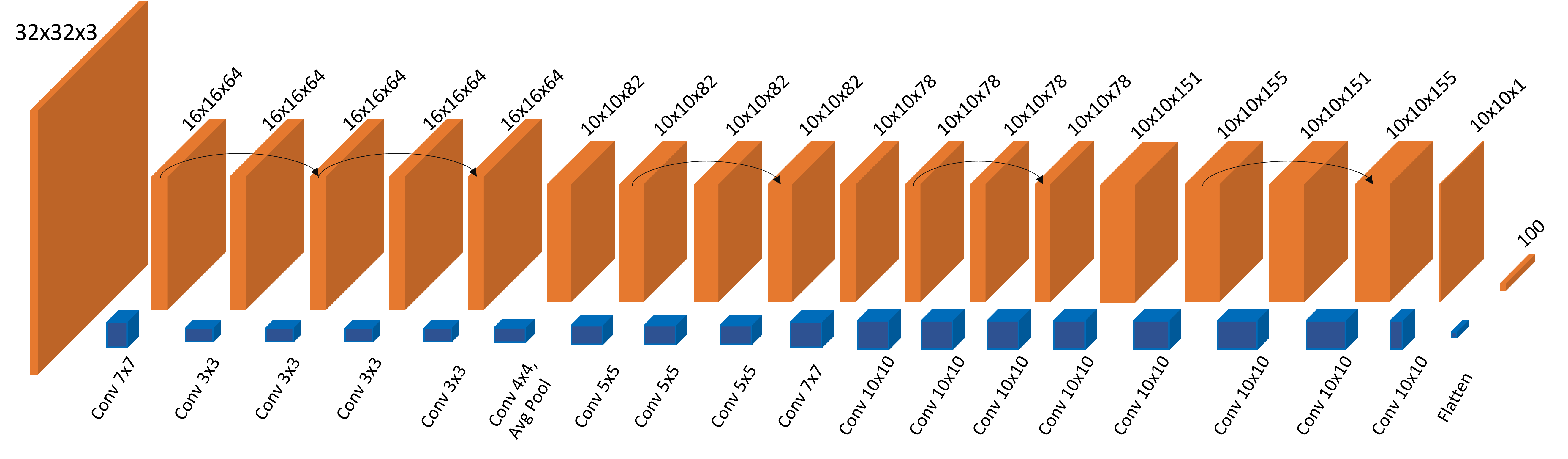}
    \end{tabular}
    \caption{Architecture comparison of our modification Resnet-18 used to train CIFAR-100 and FatNet constructed from ResNet-18 specifically for CIFAR-100 classification. (a) ResNet-18 architecture, slightly modifier from the original. Our version does not use strides since optics can not perform strides in convolutions. We also skipped the second non-residual convolutional layer to make it more compatible with CIFAR-100. (b) FatNet derived from the ResNet-18 for CIFAR-100. Compared to ResNet-18, this architecture contains fewer channels but larger resolutions. Kernel resolutions can go up to 10x10, while feature maps are not pooled lower than 10x10. The last layer is a 10x10 matrix flattened to form a vector of 100 elements, each representing a class of CIFAR-100.\label{fig:architecture}}
\end{figure}  

By having the larger feature maps and kernel sizes in the classifier CNN, we can ensure full utilization of the free-space optics. Although higher resolutions come with the problem of overfitting, our approach uses the same number of trainable parameters as the standard approach. Essentially, we have created the rules for turning any classifier into a FatNet:

\begin{enumerate}
\item The FatNet should preserve the same number of layers as the original network to keep the same number of non-linear activation functions.
\item The FatNet should keep precisely the same architecture as the original network on the shallow layers until the shape of the feature maps pools down to the shape where the number of elements of the feature map is less than or equal to the number of classes. 
\item FatNet has the same total number of pixels of the feature maps at the output of each layer as the original networks. Hence, since the feature maps' shape stays constant and does not use pooling, the new number of output channels needs to be calculated, which will be less than the original network.
\item FatNet has the same number of trainable parameters per layer as the original network. Since we have reduced the number of output channels based on the third rule, the number of trainable parameters has also been reduced. Hence new kernel size needs to be calculated based on the number of output channels. 
\end{enumerate}

It is also important to remember that the FatNet is more efficient when the number of classes is significant; for example, ImageNET contains 20,000 classes. We chose Resnet-18 as the backbone network to prove the concept and trained the network with the CIFAR-100 dataset. We chose CIFAR-100 over CIFAR-10 due to the larger number of classes and the ability to keep the feature maps in the square shape of 10x10. It is essential to know that one of the limitations of the 4f-based convolution is the failure to perform the convolutions with the stride. Since most off-the-shelf networks contain stridden convolutions, this can be a potential problem. However, we can get around the problem by replacing the stridden convolutions with the combination of standard convolution and pooling. Because we do not want to reduce the resolution of our feature maps, we decided to ignore the strides in our ResNet-18 architecture and use 2x2 MaxPooling after the first layer. 

No modification is done to the first five layers since they were all 16x16 resolution. For the following layers, we have calculated the number of pixels in each feature map and measured how many channels the layers should contain if all layers' feature maps stay 10x10. Then we calculated the number of trainable parameters in the original network(excluding bias). Based on the number of trainable parameters and the new number of channels, we have calculated the new kernel resolutions as shown in Table~\ref{tabFatNetconstruction}. 

\begin{table}[H]
\caption{Construction of FatNet from Resnet-18}
\centering
\begin{tabular}{cccc}
\toprule
\textbf{Layer}	& \textbf{Number of weights}	& \textbf{Feature pixels} & \textbf{FatNet Layer}\\
\midrule
64x128,k=(3x3)    & 73728             & 8192             & 64x82,k=(4x4)      \\
128x128,k=(3x3)   & 147456            & 8192             & 82x82,k=(5x5)      \\
128x128,k=(3x3)*2 & 147456*2          & 82192*2          & 82x82,k=(5x5) *2   \\
128x256,k=(3x3)   & 294912            & 4096             & 82x41,k=(9x9)      \\
256x256,k=(3x3)   & 294912            & 4096             & 41x41,k=(19x19)    \\
256x256,k=(3x3)*2 & 294912*2          & 4096*2           & 40,41,k=(19x19)*2  \\
256x512,k=(3x3)   & 1179648           & 2048             & 41x21,k=(37x37)    \\
512x512,k=(3x3)   & 2359296           & 2048             & 21x21,k=(73x73)    \\
512x512,k=(3x3)*2 & 2359296*2         & 2048*2           & 21x21,k=(73x73) *2 \\
FC(512,100)       & 51200             & 100              & 21x1,k=49x49       \\ \bottomrule
\end{tabular}
\label{tabFatNetconstruction}
\end{table}
\unskip

Unfortunately, kernels larger than the input features in the last layer cause a problem. The main problem is that the convolutions are the same padding type, meaning that the input and output resolutions are the same 10x10, in our case. This means the outer regions of the kernels larger than 10x10 are redundant and will not train. This restricts us to the convolutions of the kernel with a maximum resolution of 10x10. Therefore we reduced the kernel size by increasing the number of channels in those layers, which violates our third rule of the FatNet construction (see Figure \ref{fig:architecture}). However, this is the better solution since the network may underfit if the number of trainable parameters is reduced. 

Usually, the image classifier neural networks are based on the convolutional layers for the feature extraction and dense layers for the classification. Sometimes fully convolution networks end up with a convolutional layer with a 1x1 shape and the number of output channels equal to the number of classes. Our FatNet's output layer is the convolutional layer with one channel and each pixel representing the probability of the class in the classification network. In our case with CIFAR-100 training, the output shape is 10x10 with one output channel. The main advantage of FatNet and its suitability for free-space optical training is that FatNet uses fewer output channels but larger resolution feature maps and kernels. Moreover, it is a fully convolutional network, which makes it fully compatible with the 4f accelerator.

To validate our results, we also developed a simulator built on top of PyTorch as the custom layer OptConv2d. OptConv2d replaces the convolution operation of the standard convolutional layer with the simulation of the 4f inference. In order to do that, we had to simulate the propagation of the amplitude-modulated light using the Angular Spectrum of Plane Waves (ASPW) method. According to the Angular Spectrum method, if the initial wavefront is $U_1(x,y)$, the next wavefront is calculated as:
\begin{equation}
    U_2(x,y)=F^{-1}[F[U_1(x,y)] H(f_x,f_y)]
\end{equation}
where $H(f_x,f_y)$ is the transmittance function for free-space.

The transmittance function of the free-space propagation comes from the Fresnel diffraction transfer function:

\begin{equation}
    H_{F}(f_x,f_y) = \exp \Biggl[ jkz -j\pi \lambda z(f_x^2+f_y^2)\Biggr]
\end{equation}
where $k=\frac{2\pi}{\lambda}$, z is the distance travelled by light and $\lambda$ is the wavelength~\cite{li_diffraction_2007, voelz_computational_2011}.

Since the 4f system contains two lenses, the transmittance function of each lens is:

\begin{equation}
    t_{A}(x,y) = P(x, y)\exp \Biggl[ -j\frac{k}{2f}(x^2+y^2)\Biggr]
\end{equation}
where $f$ is the focal length of the lens and $P(x,y)$ is the pupil function \cite{voelz_computational_2011}.

The distance at which the angular spectrum method calculates the next wavefront depends on the pixel scale and is calculated as:
\begin{equation}
    z = \frac{N(\Delta x)^2}{\lambda}
\end{equation}
where $\Delta x$ is the pixel scale, $N$ is the number of pixels, and $\lambda$ is the wavelength. In case when the propagation distance needs to be longer than the above formula for the distance, the propagation can be calculated in several iterations. We chose such pixel scale for each propagation, so $z$ becomes equal to the focal distance of the lens. In this case, we have to do only one iteration for each focal distance propagation in the 4f system. 

The simulator uses pseudo-negativity, so each convolution is run twice to avoid the kernel's negative values in optics. Moreover, due to the laws of geometrical optics, the output of the 4f device is always rotated 180 degrees. Luckily, this is not a problem for convolutional neural networks since they can continue extracting the futures from the rotated feature maps. 

\subsection{Experiments}

The main goal of FatNet is not to gain accuracy but to demonstrate that the network with prescribed architecture can maintain accuracy by being accelerated using free-space optics and perform fewer inferences through the 4f system than the original network. Hence our experiments were aimed at testing and comparing the original network and FatNet. 

We recreated the modified version of ResNet-18, converted it to the FatNet and trained both networks. To validate the accuracy of the FatNet in the optical device, we trained the network in the simulator. In the real 4f system, we would have taken advantage of the parallelism of the network by tiling the batches. However, batches were not tiled in the simulator since the matrices are represented in PyTorch's Tensor format. All the operations were performed without unwrapping the Tensor and performed the Fourier transforms and multiplications directly on the 4-dimensional tensors. We chose this approach since the simulator-based training of the network was much slower than the standard PyTorch network. Each epoch of the optical simulation of FatNet takes 67 minutes, while the epoch in the standard FatNet with Conv2d layer of PyTorch is 15 seconds only. 

The wavelength of the laser was set to 532nm (green). We assumed the use of convex lenses with a 5mm diameter and focal distance of 10mm. It should also be noted that we have not taken the device's quantization and noise into account and used type float32. 

We split our training set into training and validation by 80-20\% ratio, respectively, making it 40,000 for training and 10,000 images for validation.
The dataset was normalised using the mean and standard derivation of the CIFAR-100 at all channels. Moreover, we have applied augmentation methods, including the horizontal flip and random crop with the padding of four. All networks were trained with the SGD optimiser, 0.9 for the momentum and with the starting learning rate of 0.01, updating every 50 steps by 0.2. The last layers of all networks were passed through the 20\% Dropout layer. We have trained all the networks with 2x NVIDIA A100 40GB GPUs.

ResNet-18 and FatNet were trained with a batch size of 64(32 per GPU). However, the optical simulation of the FatNet had to be trained with the batch size of 16 (8 per GPU) due to the high memory requirement of the simulator, as the optical simulation enhances the computational graph and number of gradients. Although we have not simulated the parallelism of the 4f system, to gain the speed-up, the 4f system needs to take advantage of high resolution. FatNet's best acceleration can be achieved if the batch tiling is performed. In order to use batch tiling, all the inputs of the same batch have to be tiled in one input block, and the kernel has to be padded to the same size as the input block. Before tiling the inputs, they must be individually padded to $M+N-1$, where $M$x$M$ is the input size, and $N$x$N$ is the kernel size. According to this method, the number of possible batch sizes can be calculated as follows: 

\begin{equation}
    n = \lfloor\frac{R}{M+N-1}\rfloor^2
\end{equation}
where $R$ is the resolution of the 4f system and $\lfloor \cdot \rfloor$ is the floor function.

\section{Results}

\begin{table}[H]
\caption{ Comparison of the test accuracy and number of convolution operations used in each experimented network}
\begin{tabular}{llll}
\toprule
\textbf{Architecture}	& \textbf{Test Accuracy}	& \textbf{Number of Conv Operations} & \textbf{Number of Conv Operations}\\ \\
&  \small{$mean\pm std$} & & Ratio to baseline\\
\midrule
ResNet-18	                 	& $66 \pm 1.4\%	$		& 1,220,800      &  1(baseline)\\
FatNet	                     	& $60 \pm 1.4\%	$		& 148,637        &  0.12\\
Optical simulation FatNet		& 60\%			        & 148,637        &  0.12\\
\bottomrule
\end{tabular}
\label{tab1}
\end{table}
\unskip

Based on the configurations described above, our implementation of ResNet-18 achieved an accuracy of 66\%. In contrast, FatNet's implementations, both with GPU and simulation of optics lagged in accuracy, resulting in 60\%. However, FatNet implementation performs 8.2 times fewer convolution operations to reach this level of accuracy and does not require any dense layers for classification. 

\begin{figure}[H]
\centering
\includegraphics[width=15cm]{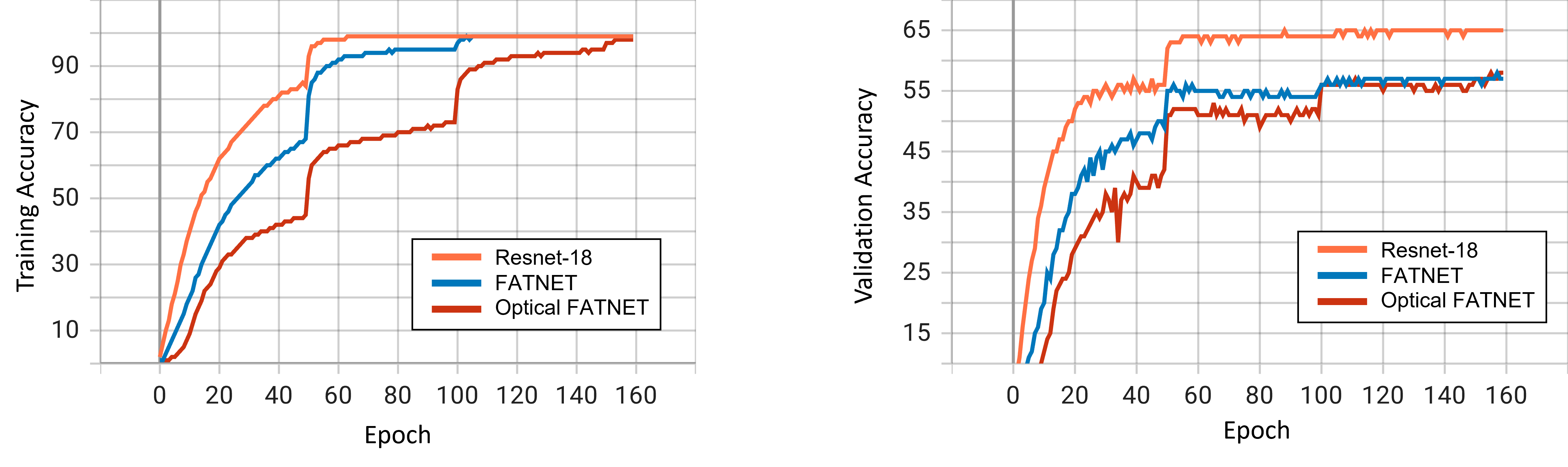}
\caption{Training and validation accuracy for each experimented network at every epoch\label{fig:acc}. (a) Training accuracy of ResNet-18, FatNet and Optical simulation of FatNet. All networks achieved an accuracy of 99\%. However, the ResNet-18 required fewer epochs. On the other hand, the optical simulation took longer to train since it uses a more extended computation graph to simulate light propagation. (b)  Validation accuracy of ResNet-18, FatNet and Optical simulation of FatNet. ResNet-19 trained up to 66\%, while FatNet could not achieve reach validation and test accuracy higher than 60\%, although it performed fewer convolution operations.}
\end{figure} 

The same can be said about the training process. Since it may take more epochs for the FatNet to reach the desired accuracy, this architecture is only beneficial if accelerated with the 4f system. 

The measured and calculated inference time for FatNet and ResNet-18 with optics and GPU were obtained and observed. The observations were conducted based on the batch size of 64, like our experiments, and 3136 maximum utilization of 4f system with 4k resolution modulators and camera. 

\begin{table}[H]
\centering
\caption{ Inference time in seconds per input for ResNet-18 and FatNet with optics and GPU with a batch size of 64 and 3136 for cases when the 4k resolution of the 4f device is fully utilised. The frame rate of the 4f device is approximated at 2MHz\cite{li_channel_2020}.}
\begin{tabular}{ccc}
\toprule
\textbf{Architecture}	& \textbf{Batch 64}	& \textbf{Batch 3136} \\
\midrule
ResNet-18(GPU)	                 	& $1.350\mathrm{e}{-4}$		& $1.167\mathrm{e}{-4}$\\
FatNet(GPU)	                     	& $4.565\mathrm{e}{-4}$		& $7.942\mathrm{e}{-4}$\\
ResNet-18(Optics)	            	& $3.815\mathrm{e}{-2}$	    & $7.786\mathrm{e}{-4}$\\
FatNet(Optics)	                	& $4.645\mathrm{e}{-3}$	    & $9.479\mathrm{e}{-5}$\\
\bottomrule
\end{tabular}
 \label{tab3}
\end{table}
\unskip

\section{Discussion}

Although the FatNet does not converge as well as the ResNet-18, it is still 8.2 times faster, if both were to be trained with optics. CIFAR-100 is an extended dataset of CIFAR-10, but unlike CIFAR-10, CIFAR-100 is much harder to train. Numerous researchers have tried different augmentation and regularisation methods to improve the performance of the classification the CIFAR-100. For instance, Mizusawa~\cite{mizusawa_interlayer_2021} tried the interlayer regularization method and improved the accuracy of the classification of CIFAR-100 in ResNet-20 from an average of 64.09\% to 65.59\%. Shah~\cite{shah_deep_2016} used ELU activation layers to improve the CIFAR-100 accuracy from 72.77\% to 73.45\%. Our modification of ResNet-18 reached the average test accuracy of 66\%, which is comparable to Mizusawa but lower than Shah. Then, our tests of FatNet showed that by sacrificing only 6\% of test accuracy, we could perform 8.3 times fewer convolutions in optics, which will mean fewer conversions from optics to electronics and vice versa. 

The training accuracy graph in Figure \ref{fig:acc} shows that the network trained with the optical simulation trains slower than in other experiments. When simulating the 4f system, PyTorch was using the simulation of light propagation as part of the computation graph of the neural network, which vastly increased the computation graph. This caused a slowdown in the network training. From the point of view of validation accuracy, the FatNet trained with the GPU, and its optical simulation does not alter much, especially after the first learning rate step on epoch 50. Although the validation accuracy of FatNet and optical simulation of FatNet did not exceed 57\% and 58\%, respectively, the test accuracy reached 60\% in both cases. This difference is caused by the augmentation applied only to the validation and training sets and not to the test set.

However, it should be noted that the acceleration in a 4f system with FatNet is only possible if the parallelism of the 4f system is utilized not with the channel or kernel tiling but with the batch tiling. The increase in resolutions and reduction of the number of channels will not change the performance much if channel tiling is used. Unfortunately, due to the high latency of the modern light modulators and cameras, it is almost impossible to get an acceptable acceleration over GPU with 4f, with the efficiency batch size as shown in Table~\ref{tab3}. However, the 4f system's acceleration is almost equalised to the GPU compared to the non-GPU inference (see Table~\ref{tab3}). If we fully utilise the 4K resolution of the 4f system, the batch size of 3136 can be used, and the acceleration of the 4f system over GPU becomes obvious. Moreover, it can be seen that the use of FatNet improves the speed of the inference in optics and works completely opposite way with the GPU, regardless of the batch size. However, enormous batch sizes like this are not efficient and will lead to overfitting.

Moreover, it should be mentioned that in our experiments, we have not tested the network with different quantization levels and noise which can occur in the system. Low precision training can potentially affect the test accuracy of the network, but there have been many successful attempts to train the neural networks with low precision to save on memory or accelerate the inference. On the other hand, noise can be beneficial and can be used as a regularisation method. Since random and unpredictable noise can be a sort of augmentation method for our dataset. If we use a smaller bit depth, the noise may not affect the accuracy since the changes in resulting light intensity will be low.

Another issue that is important to consider is the alignment of the optical
elements. One of the main disadvantages of the 4f system compared to the silicon photonics approach is the alignment of the optical elements. A slight alteration in the alignment of the elements of the 4f system can lead to entirely wrong results and to the inability to keep track of the graph correctly. Unfortunately, our simulator is not designed to consider the alignment problems. In practice, optical cage systems can be used to keep the elements fixed and aligned. 

The design of the FatNet makes it more suitable for the datasets with a large number of classes, like 100 in our case, but also it can potentially work with images of a higher resolution. Unfortunately, the simulation of the light propagation takes a large amount of the GPU memory, which is why we did not use ImageNet for our experiments when it was the most obvious choice for the FatNet. 

\section{Conclusions}

In this research, we looked at a new way of fully utilizing the high-resolution capabilities of the 4f system for classification. We introduced the transformation method, which makes the regular neural network designed for the CPU/GPU training more compatible with the free-space optical device. After testing the FatNet with the CIFAR-100 dataset, using the ResNet-18 as the backbone network and the optical simulation of the FatNet using the angular spectrum method, we reached a test accuracy of 66\% with ResNet and 60\% with FatNet. Eventually, we demonstrated that FatNet performs 8.2 times fewer convolution operations than ResNet-18 without losing frame rate when both were implemented in optics. Compared to the standard ResNet-18, FatNet is always faster than ResNet-18 run with the optical device and also than ResNet-18 ran with GPU when the batch size is as large as 3136. Moreover, our research demonstrates the importance of using high-resolution kernels in CNN, especially in the future when the speed of cameras and light modulators improves.

\paragraph{Author Contributions:}
Conceptualization, R.I.; methodology, R.I.; software, R.I.; validation, R.I., T.C. and C.C.R.-A.; formal analysis, R.I.; investigation, R.I.; resources, R.I.; writing---original draft preparation, R.I.; writing---review and editing, C.C.R.-A. and T.C.; visualization, R.I.; supervision, T.C. and C.C.R.-A.; Y.Y. All authors have read and agreed to the published version of the manuscript.

\paragraph{Funding:}
This research received no external funding
\paragraph{Data Availability Statement:\\}
Simulator and the models repository: \url{https://github.com/riadibadulla/simulator}\\
CIFAR-100 dataset: \url{https://www.cs.toronto.edu/~kriz/cifar.html}
\paragraph{Conflict of interest:}
The authors declare no conflict of interest.

\paragraph{Abbreviations:\\}
The following abbreviations are used in this manuscript:\\
\noindent 
\begin{tabular}{@{}ll}
CIFAR & Canadian Institute For Advanced Research\\
CNN & Convolutional Neural Network\\
ASIC & Application-Specific Integrated Circuit\\
ELU & Exponential Linear Unit\\
SGD & Stochastic Gradient Descent \\
FFT & Fast Fourier Transfer\\
TPU & Tensor Processing Unit\\
MZI & Mach–Zehnder Interferometer\\
SLM & Spatial Light Modulators
\end{tabular}


\bibliographystyle{IEEEtranN}
\bibliography{references}  

\end{document}